\documentclass{article}
\usepackage{spconf,amsmath,amsfonts,graphicx}
\usepackage{multicol,multirow,color}
\usepackage{mathtools}
\usepackage{colortbl}
\usepackage{booktabs}
\usepackage{tabularx}
\usepackage{balance}
\usepackage{url}
\usepackage[hidelinks]{hyperref}
\usepackage[labelfont=bf]{caption}
\title{On the detection of synthetic images generated by diffusion models}
\name{Riccardo Corvi\textsuperscript{$\star$}, Davide Cozzolino \textsuperscript{$\star$},  Giada Zingarini\textsuperscript{$\star$}, Giovanni Poggi\textsuperscript{$\star$},  Koki Nagano\textsuperscript{$\dagger$}, Luisa Verdoliva\textsuperscript{$\star$} 
}
\address{$\star$ University Federico II of Naples \\ $\dagger$ NVIDIA}

\begin{document}
\maketitle
\begin{abstract}
Over the past decade, there has been tremendous progress in creating synthetic media,
mainly thanks to the development of powerful methods based on generative adversarial networks (GAN).
Very recently, methods based on diffusion models (DM) have been gaining the spotlight.
In addition to providing an impressive level of photorealism, they enable the creation of text-based visual content,
opening up new and exciting opportunities in many different application fields, from arts to video games.
On the other hand, this property is an additional asset in the hands of malicious users,
who can generate and distribute fake media perfectly adapted to their attacks,
posing new challenges to the media forensic community.
With this work, 
we seek to understand how difficult it is to distinguish synthetic images generated by diffusion models from pristine ones and whether current state-of-the-art detectors are suitable for the task.
To this end, first we expose the forensics traces left by diffusion models, 
then study how current detectors, developed for GAN-generated images, perform on these new synthetic images,
especially in challenging social-networks scenarios involving image compression and resizing.
Datasets and code are available at 
\href{https://github.com/grip-unina/DMimageDetection}{\url{github.com/grip-unina/DMimageDetection}}.
\end{abstract}
\section{Introduction}
\label{sec:intro}

The use of diffusion models for the generation of synthetic media is arousing great interest
among both researchers and practitioners.
Besides the high quality and photorealism of the generated images, 
it is the opportunity to model a wide variety of subjects and contexts that appears to be particularly interesting.
In fact, diffusion models can be guided by textual descriptions or pilot sketches 
to generate images of a virtually unlimited set of categories, bounded only by our imagination.
Therefore,
this technology represents a powerful tool for artists, game designers, and any type of creative users.
Unfortunately, this includes also malicious users,
that may take advantage of this increased flexibility to generate fake media more fit to their disinformation goals.
In this paper, we try to assess how prepared we are to face this new threat.

Some recent papers have begun studying the detection DM-based images. 
In particular, in \cite{Farid2022lighting, Farid2022perspective} it was noted that
the lack of explicit 3D modeling of objects and surfaces causes asymmetries in shadows and reflected images.
Furthermore, global semantic inconsistency can be observed, to some extent, in lighting.
These traces can certainly be exploited to identify today's DM images.
However, if the rapid advancement of GAN images can be taken as a paradigm,
new DM-based methods can be expected to soon overcome these limitations
and generate images that satisfy all necessary semantic constraints, be it lighting, perspective or any other aspect.

\begin{figure}[t!]
    \centering
    \includegraphics[page=1, width=1.45\linewidth, trim=20 20 0 0]{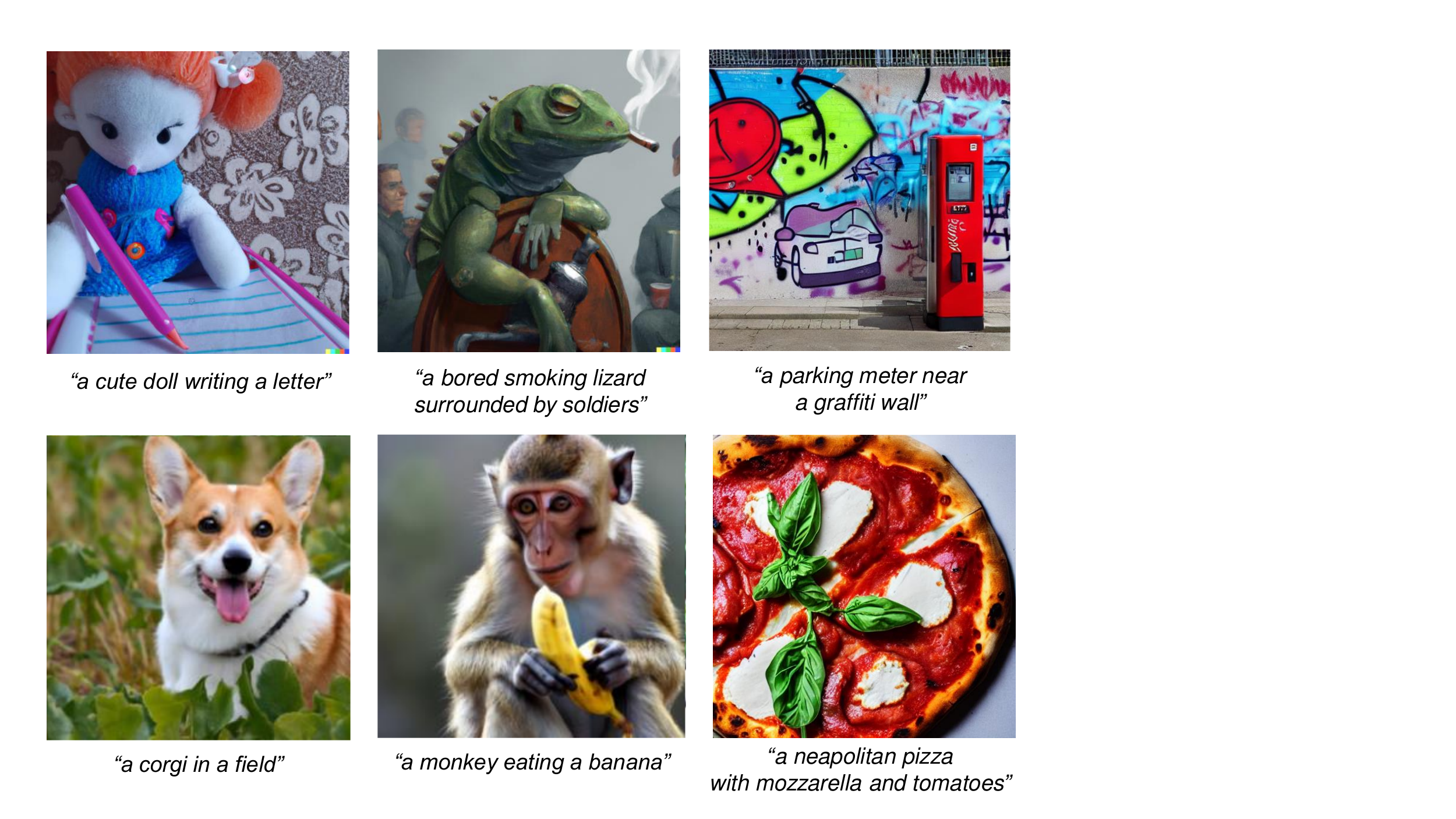}
    \caption{\small Synthetic images generated using recent text-to-image models: DALL·E 2 \cite{ramesh2022hierarchical}, stable diffusion \cite{stablediffusion2022} and GLIDE \cite{nichol2021glide}.}
    \label{fig:synthetic_images}
\end{figure}

Indeed, most state-of-the-art detectors of fake images rely on traces that are invisible to the human eye.
Even images that are visually perfect, with no evident semantic inconsistency, 
can be distinguished from real images based on the traces that are inherent of the generation process.
In fact, any method used to create synthetic visual data 
embeds some peculiar traces in their output images, that are related to the actions taken in the generation process.
These traces are different from those typical of modern digital devices 
enabling fake image detection \cite{Verdoliva2020media}.
Moreover, 
each generation architecture is characterized by its own peculiar traces, thereby allowing also for source attribution.
The presence and distinctiveness of such traces has been proven
by extracting a sort of spatial-domain artificial fingerprints \cite{Marra2019DoGAN, Yu2019}, 
but also through frequency-domain analyses
showing that the upsampling operation performed in most GAN architectures give rise to clear spectral peaks \cite{Zhang2019, frank2020leveraging}.

Based on these concepts, several promising CNN-based detectors of GAN images have been developed.
However, the problem is far from being solved.
On one hand, new and more sophisticated generation architectures are proposed by the day, some of them, like StyleGAN3 \cite{karras2021alias}, aiming explicitly at minimizing the presence of these undesired traces.
On the other hand, even state-of-the-art detectors, based on a supervised learning approach, 
have a hard time generalizing to architectures never seen in training.
Moreover, they suffer a significant performance drop when image quality is impaired, 
as it always happens on social networks, which routinely apply some resizing and compression operations.
This is because forensics traces are very weak and can be easily removed even by these non-malicious processing steps.

An interesting experiment on generalization has been recently conducted by NVIDIA within the DARPA SemaFor program.
Performers were asked to detect StyleGAN3 images, 
with the constraint that no images generated with this architecture could be used in training.
Despite the inherent difficulty of the task, some encouraging results have been achieved \cite{Nagano2021link}.
A more recent contest (IEEE VIP Cup) extended this analysis to images coming also from diffusion models architectures, with encouraging results \cite{vipcuplink}.
Also in \cite{Sha2022fake} it is proposed an initial study on the detection and attribution of diffusion models that looks promising. However, results are presented only in ideal conditions and no robustness analysis is considered.

This work aims at providing some more information on the detection of DM images and, possibly, 
contributing some useful guidelines for further developments.
In particular, we want to answer two fundamental questions:
{\it i)}  
are DM images characterized by hidden artifacts similar to those observes in GAN images? 
{\it ii)} 
To what extent are current state-of-the-art detectors effective on this type of images? 
To answer these questions, we generated a large variety of synthetic images using the most recent generators.
Then, we carried out an analysis of their artifacts, 
and finally studied the performance of some deep learning-based detectors on them,
not only in ideal conditions but also in more challenging scenarios.
where images are compressed and resized.

\section{Background}
\label{sec:backgr}

In this section we summarize the most important findings in the literature
towards the development of successful and robust deep learning-based detectors for synthetic images.
Much of the previous research relates to images generated by GANs as these architectures have so far dominated the field.

A widely agreed fact is the key importance of augmentation, including especially blurring and compression, to ensure robustness.
On the same line, training set diversity was found to help generalizing to unseen architectures, 
as shown in \cite{Wang2020} where a simple pre-trained Resnet50 is trained on 20 different categories of ProGAN images.
Such observations have been reinforced by later works \cite{Gragnaniello2021GAN, Mandelli2022detecting} 
proving the tight relation between augmentation and training diversity, on one side, 
and detectors' reliability, on the other.
Working on local patches also appears to be important \cite{chai2020makes} 
as well as analyzing both local and global features \cite{Ju2022fusing}.

Another central discovery concerns the need to avoid any loss of information 
in the pre-processing of training and test images, as well as in all layers of the neural network,
especially those closest to the input.
Most of all, it is important to avoid resizing, a common practice in deep learning to adapt images to fixed input layers,
as this entails image resampling and interpolation,
which may erase the subtle high-frequency traces left by the generation process.
To preserve these precious (and invisible) forensics artifacts, several strategies can be considered: 
{\em   i)} training the networks on local patches, cropped from the image with no resizing;
{\em  ii)} making the final decision on the whole image through some fusion strategy; 
{\em iii)} avoid downsampling steps in the first layers of the network \cite{Gragnaniello2021GAN}, as also suggested in related forensics applications.

These measures allow to minimize information losses in the precious high-frequency image components (see also Section 3).
A more extensive analysis of the performance of synthetic image detectors \cite{Gragnaniello2021GAN}
shows that pre-training all models on large datasets (e.g., ImageNet) keeps being important,
while using residuals in input instead of the original images does not improve performance,
and extreme augmentation provides only marginal gains.

\begin{figure*}[t!]
        \centering
	\begin{tabular}{c@{\hspace{1mm}}c@{\hspace{1mm}}c@{\hspace{1mm}}c@{\hspace{1mm}}c}
         ProGAN &  BigGAN & StyleGAN2 &  Taming Tran. &  DALL·E Mini   \\
	    \includegraphics[width=0.19\linewidth]{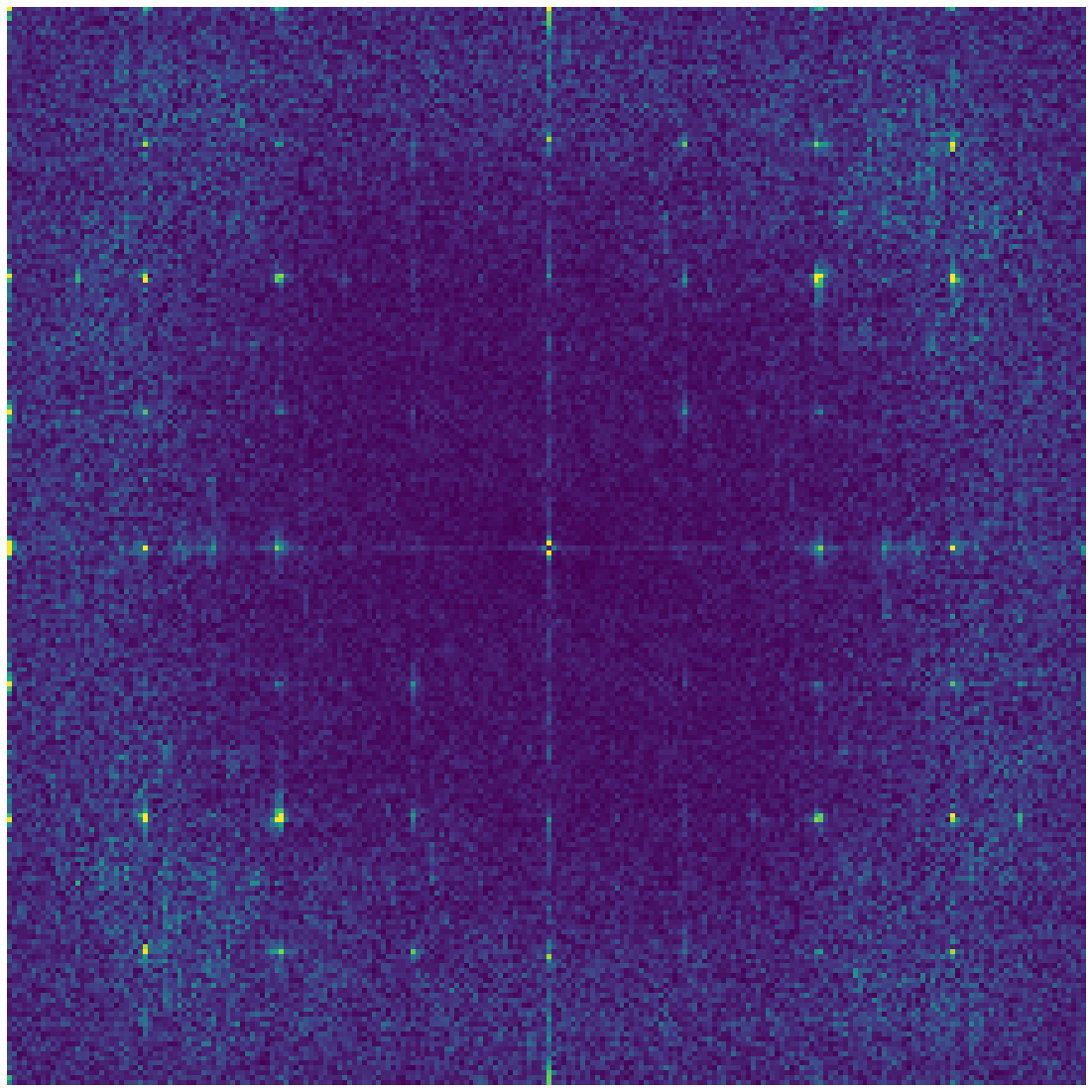} &
	    \includegraphics[width=0.19\linewidth]{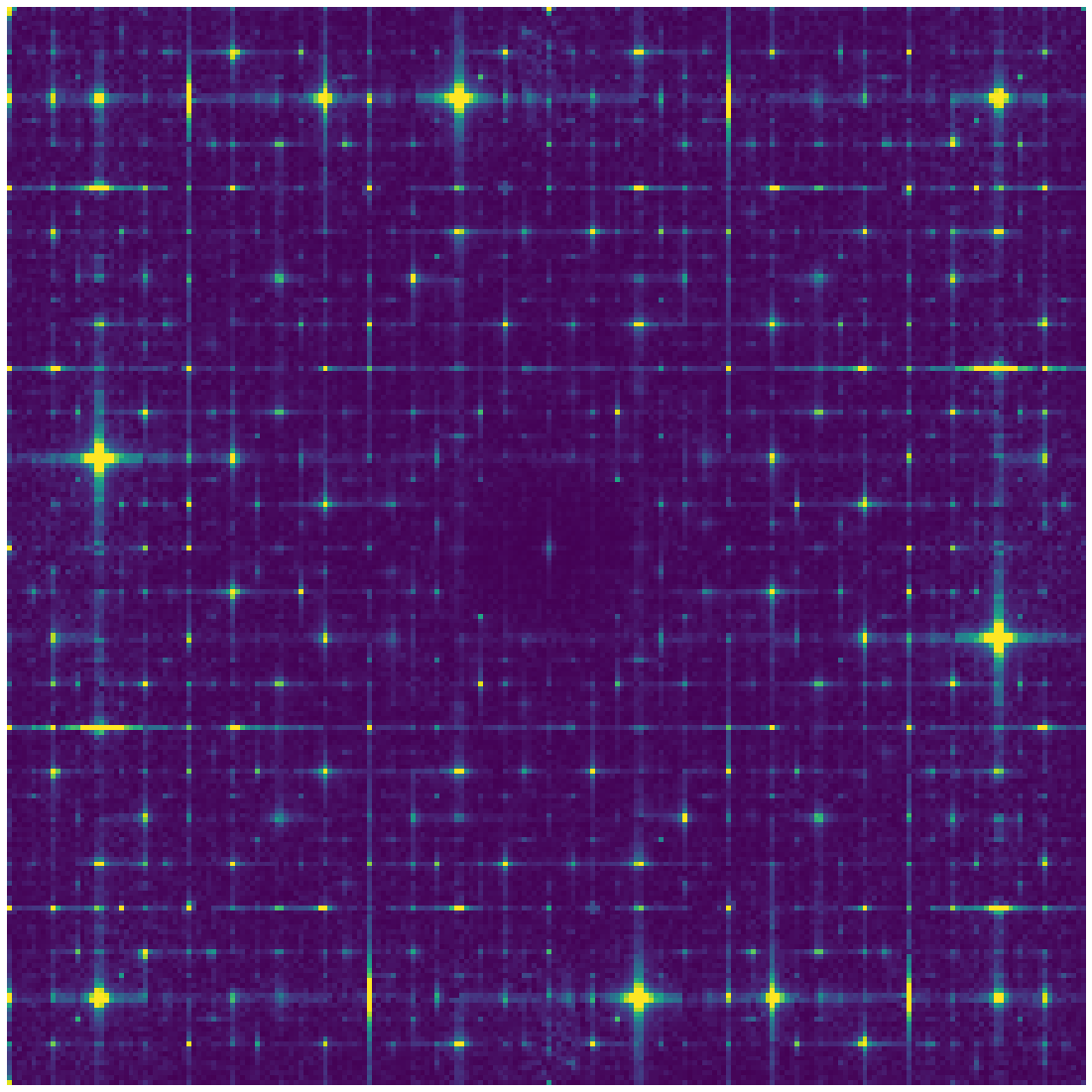} &
	    \includegraphics[width=0.19\linewidth]{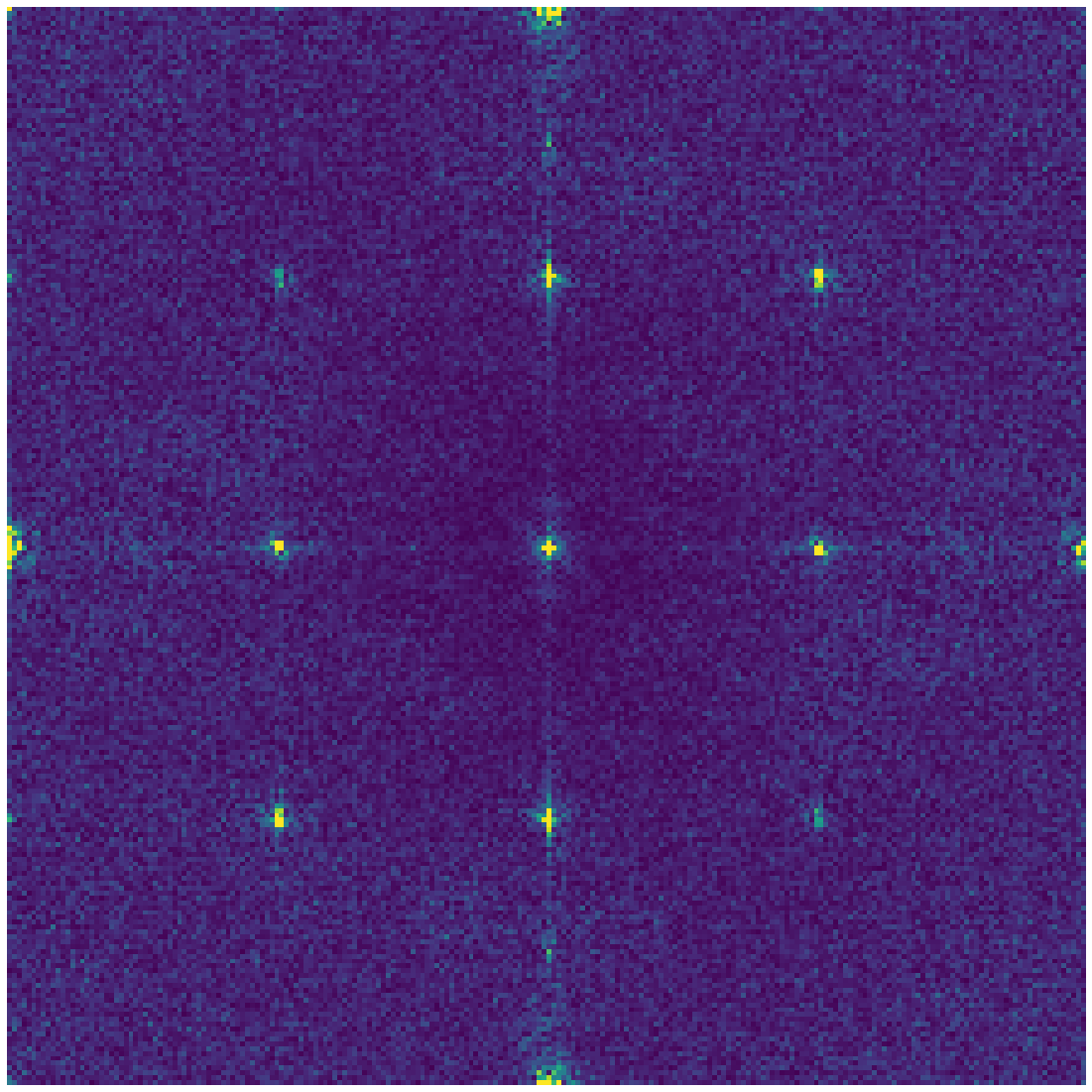} &
	    \includegraphics[width=0.19\linewidth]{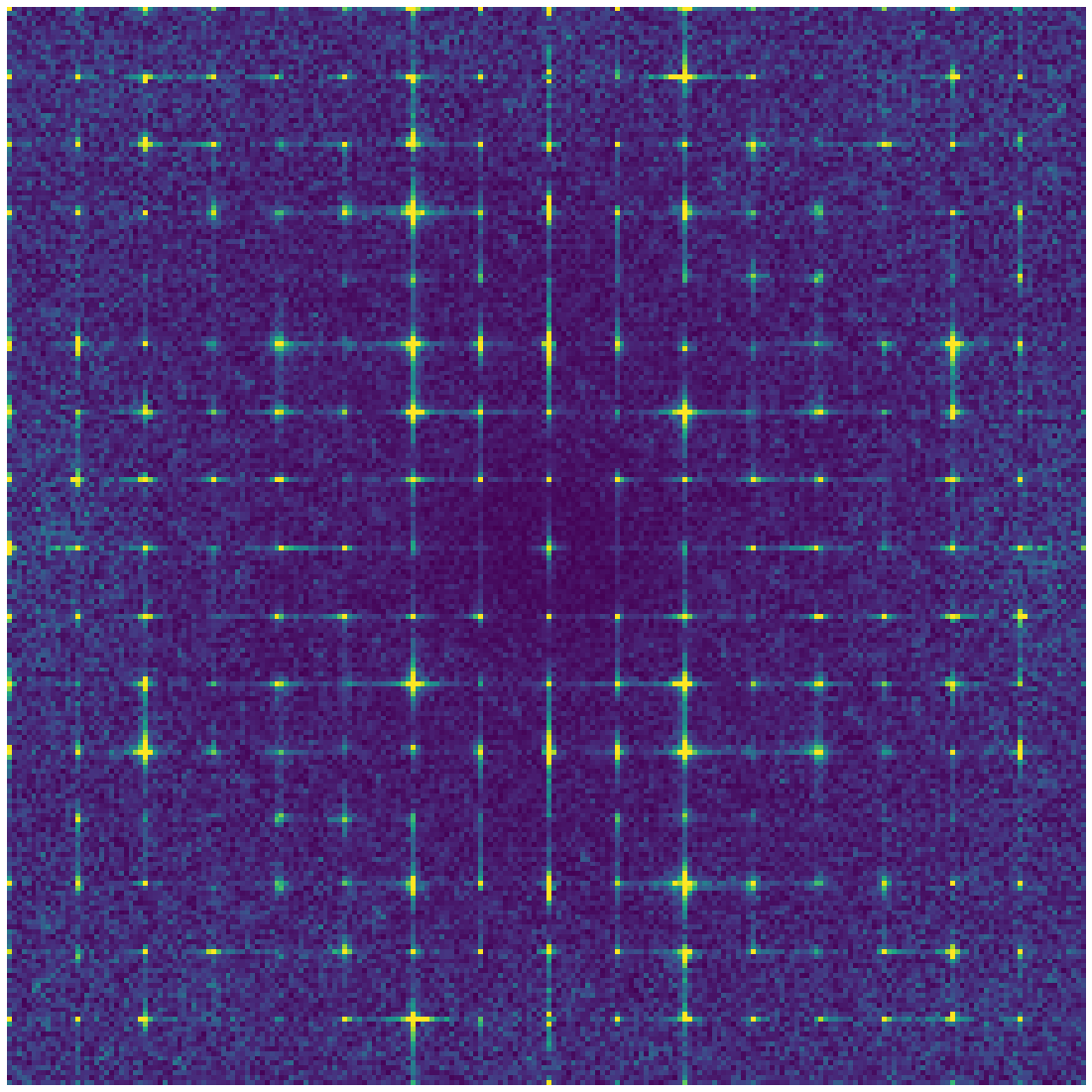} &
	    \includegraphics[width=0.19\linewidth]{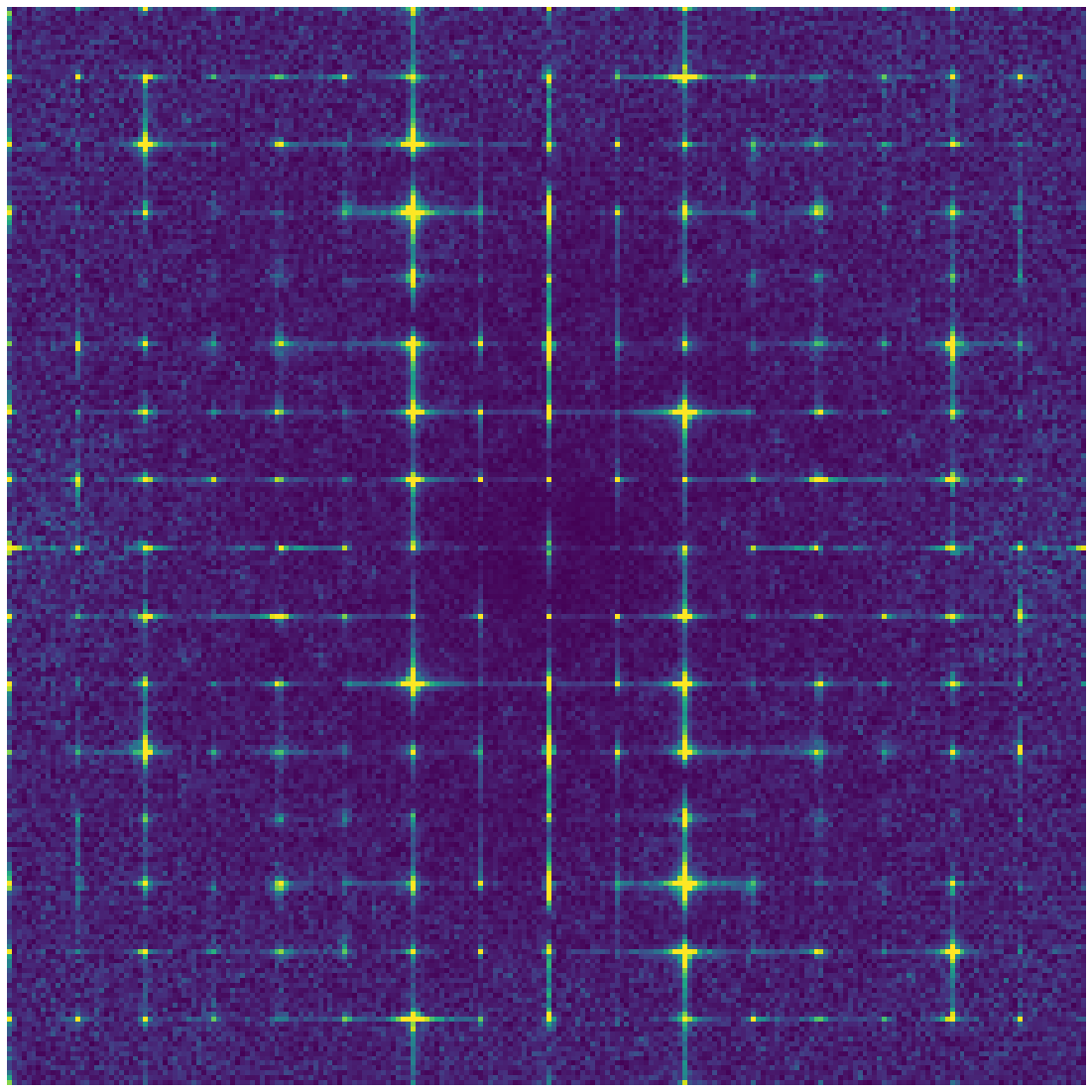} \\
	     GLIDE   & Latent Diffusion &  Stable Diffusion & ADM & DALL·E 2  \\
	    \includegraphics[width=0.19\linewidth]{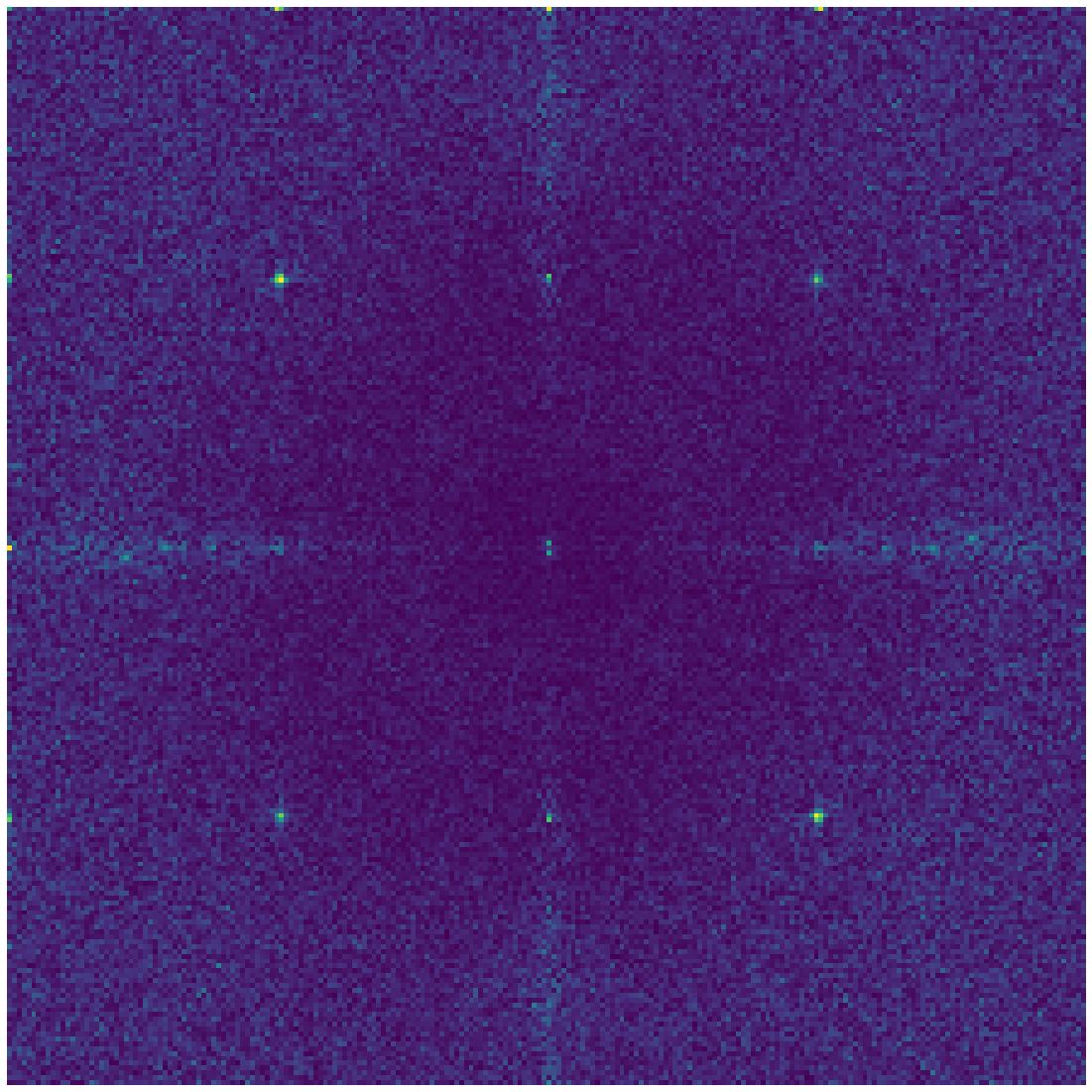} &
	    \includegraphics[width=0.19\linewidth]{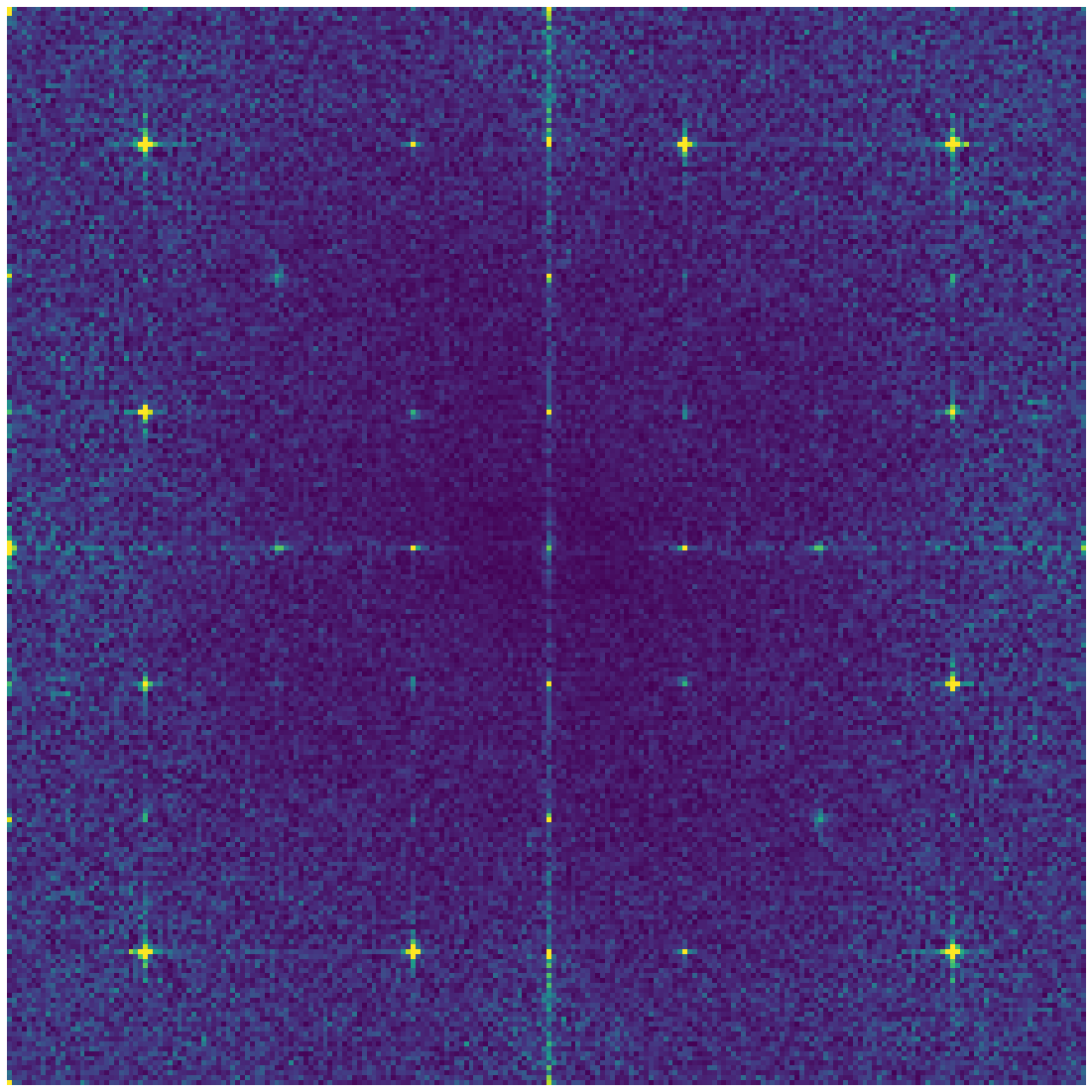} &
		\includegraphics[width=0.19\linewidth]{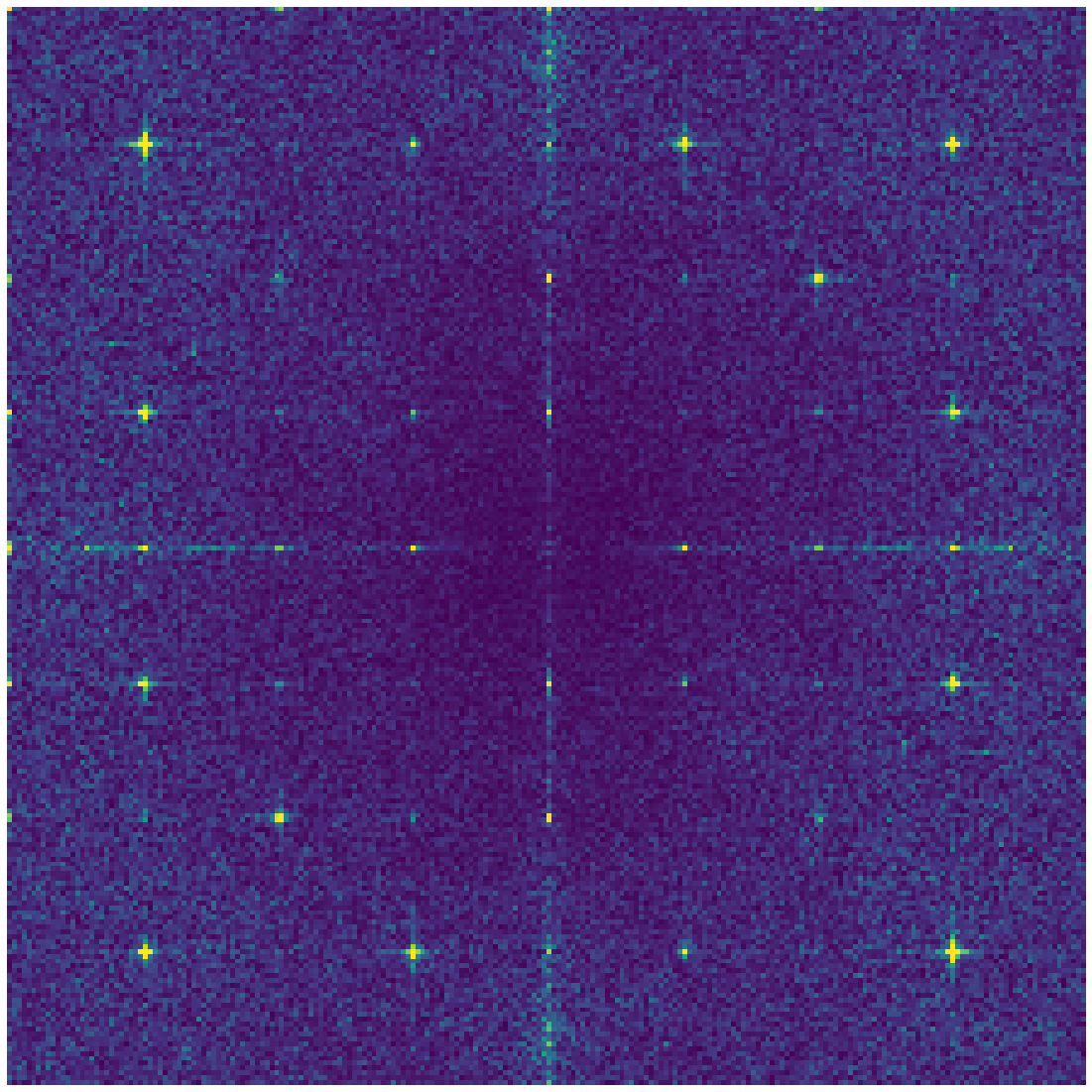} &
		\includegraphics[width=0.19\linewidth]{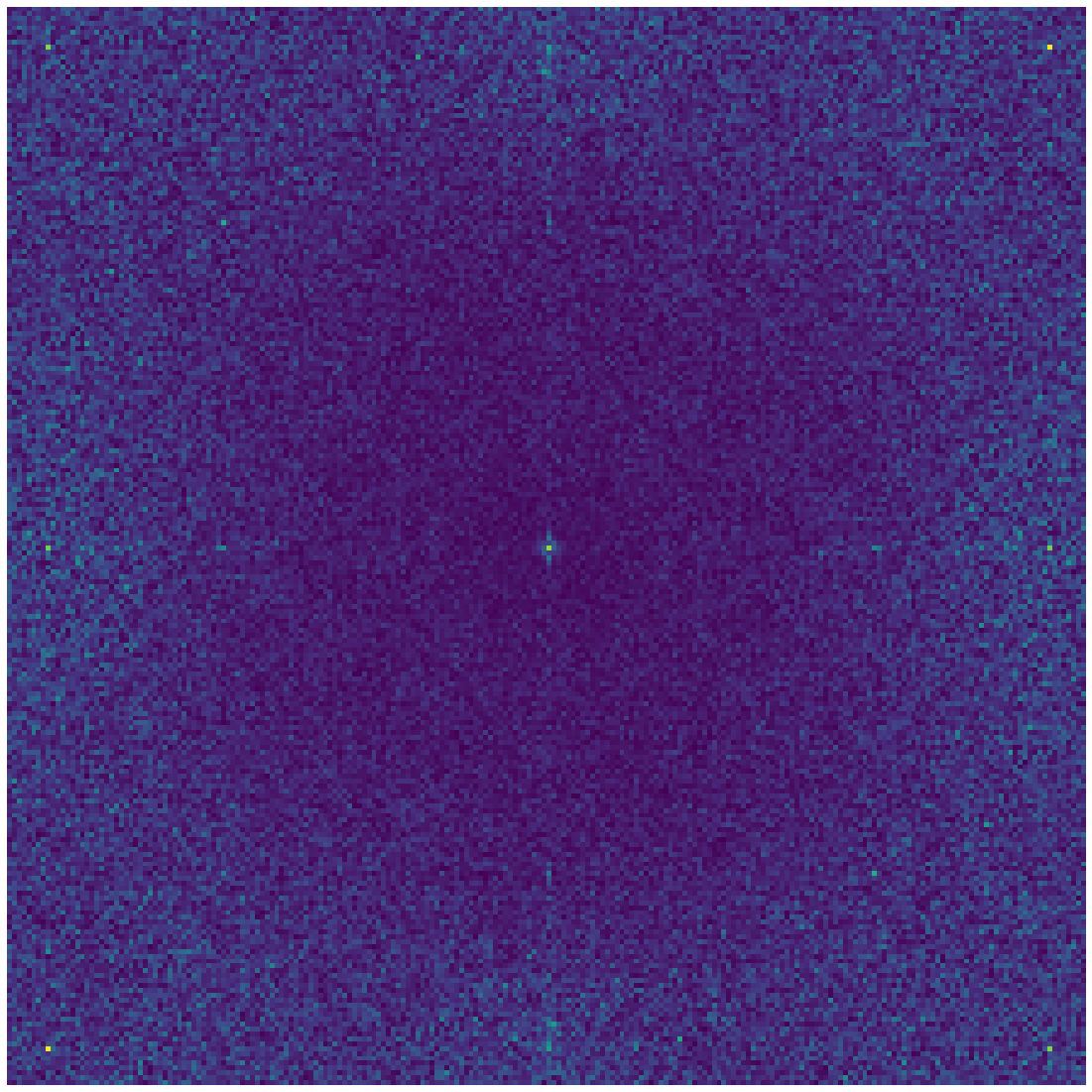} &
        \includegraphics[width=0.19\linewidth]{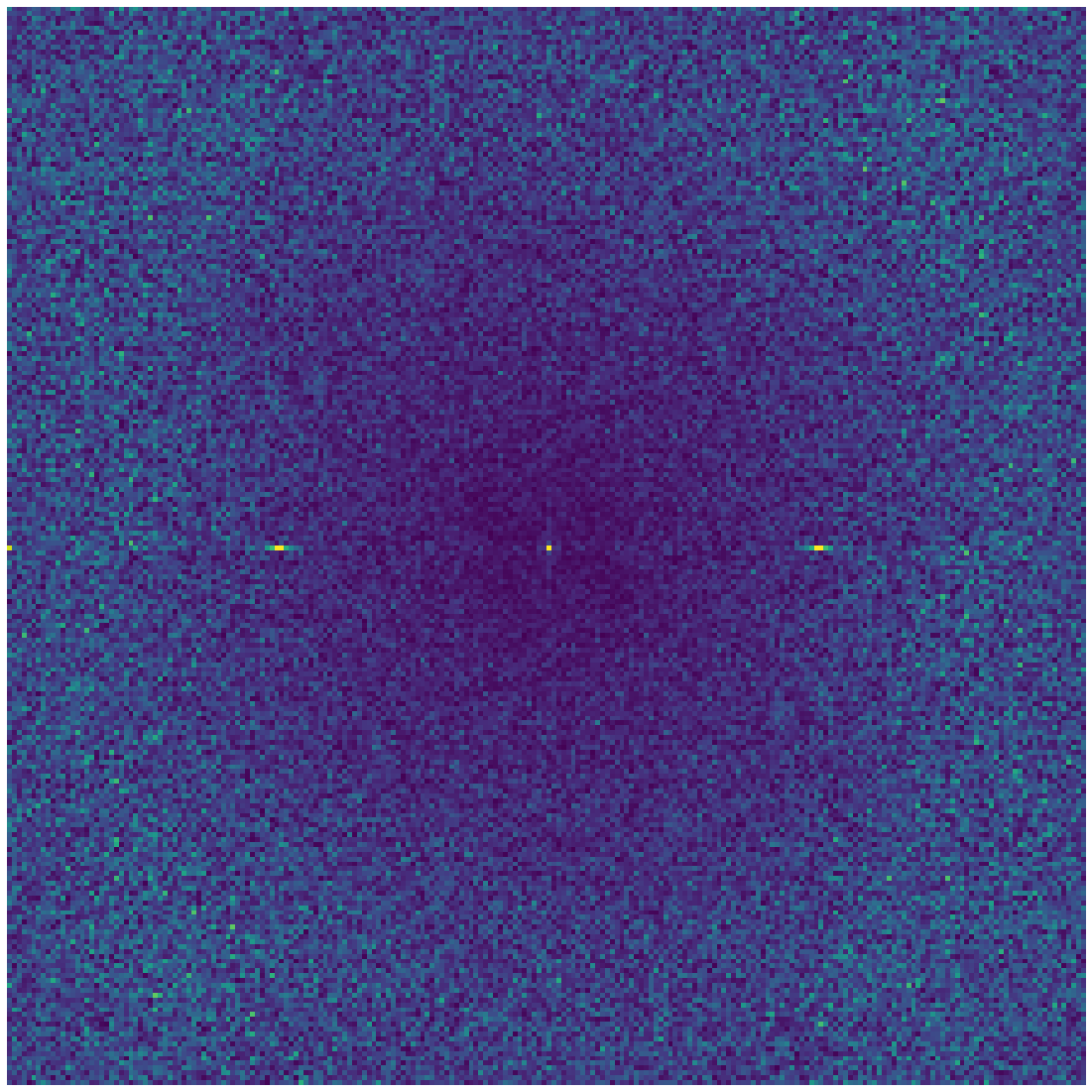} \\
        \end{tabular}
	\caption{\small Fourier transform (amplitude) of the artificial fingerprint estimated from 1000 image residuals. Top row: from left to right ProGAN \cite{karras2018progressive}, BigGan \cite{brock2018large}, StyleGAN2 \cite{karras2020analyzing}, Taming Transformers \cite{esser2021taming}, DALL·E Mini \cite{Dayma_DALLE_Mini_2021}. Bottom row: 
	GLIDE \cite{nichol2021glide}, Latent Diffusion \cite{rombach2022high}, Stable Diffusion \cite{stablediffusion2022}, ADM \cite{dhariwal2021diffusion}, DALL·E 2 \cite{ramesh2022hierarchical}
	}.
	\label{fig:GAN_fingerprints}
\end{figure*}

\section{Artifact analysis}
\label{sec:art}

Previous work \cite{Marra2019DoGAN, Yu2019} established the existence of GAN fingerprints
and their dependence on both the GAN architecture (number and type of layers) and its specific parameters (filter weights).
In particular, in \cite{Marra2019DoGAN}, fingerprints are extracted in the spatial domain
mimicking the pipeline used to extract the PRNU pattern for device identification \cite{Lukas2006}.
First, the scene content is estimated through a denoising filter $f(\cdot)$, $\widehat{X}_i=f(X_i)$,
and removed from the input image to obtain the so-called noise residual, $R_i = X_i-f(X_i)$.
This latter is assumed to be
the sum of a deterministic component, the GAN fingerprint $F$, and a random noise component $W_i$,
so the fingerprint is estimated by simply averaging a large number residuals $\widehat{F} = (1/N) \sum_{i=1}^N R_i$.

In this work we use the same procedure,
using the denoising filter proposed in \cite{Zhang2017beyond},
which proved already successful for camera fingerprint extraction \cite{Cozzolino2020noiseprint}.
We average the noise residuals of 1000 images,
then take the Fourier transform of the result to carry out a spectral analysis.
Fig.\ref{fig:GAN_fingerprints} shows the amplitude of these spectra for several architectures of interest,
both GANs \cite{karras2018progressive,brock2018large,karras2020analyzing} or VQ-GANs \cite{esser2021taming,Dayma_DALLE_Mini_2021} (top row),
and DMs \cite{nichol2021glide,rombach2022high,stablediffusion2022,ramesh2022hierarchical} (bottom row).
For all GANs, strong peaks are clearly visible in the spectra \cite{Zhang2019, Wang2020},
implying the presence of quasi-periodical patterns, the fingerprints, in the synthetic images.
Interestingly,
the same happens with some recent diffusion models, such as GLIDE, Latent Diffusion and Stable Diffusion,
which suggests good results for fingerprint-based forensic tools also in these cases.
On the other hand, such peaks are much weaker for other architectures ADM and DALL·E 2,
predicting more controversial results in these cases, as will be confirmed by the experimental analysis on next Section.

\section{Detection performance}
\label{sec:typestyle}

In this Section we present the results of experiments carried out on 
images generated by several state-of-the-art generative models including GANs, transformers, and DMs:
ProGAN \cite{karras2018progressive},
StyleGAN2 \cite{karras2020analyzing}, 
StyleGAN3 \cite{karras2021alias}, 
BigGAN \cite{brock2018large}, 
EG3D \cite{chan2022efficient}, 
Taming Transformer \cite{esser2021taming}, 
DALL·E Mini \cite{Dayma_DALLE_Mini_2021},
DALL·E 2 \cite{ramesh2022hierarchical}, 
GLIDE \cite{nichol2021glide}, 
Latent Diffusion \cite{rombach2022high}, 
Stable Diffusion \cite{stablediffusion2022} and ADM (Ablated Diffusion Model) \cite{dhariwal2021diffusion}. 
For text-to-image data we use language prompts from COCO validation and training set. 
Real data instead come from COCO \cite{lin2014microsoft}, ImageNet \cite{deng2009imagenet} and UCID \cite{schaefer2003ucid}.

We are especially interested in the ability of detectors to generalize to unseen architectures, relevant for realistic scenarios.
Therefore, we train on images generated by a single model, and test on all others.
In detail, we consider two cases: 
a) training only on ProGAN images, using the same setting as in \cite{Wang2020} (362K fake images with 20 categories);
b) training only on images generated by Latent Diffusion (200K fake images with 5 categories).
In both cases, tests are performed on 1000 synthetic images for each model and 5000 real images.

We compare the following detectors: 
Spec \cite{Zhang2019}, based on frequency analysis, 
PatchForensics \cite{chai2020makes}, that relies on local patch analysis, 
Wang2020 \cite{Wang2020}, a ResNet50 with blurring and compression augmentation, 
and Grag2021 \cite{Gragnaniello2021GAN}, same backbone as before but avoiding down-sampling in the first layer and intense augmentation. Results are given in terms of area under the receiver-operating curve (AUC) and accuracy at the fixed threshold of 0.5.

\begin{table*}
    \centering
    {\footnotesize
    \begin{tabular}{lcccccccc}
    \toprule 
    & \multicolumn{8}{c}{Trained on ProGAN}  \\
 \cmidrule(lr){2-9}
    & \multicolumn{4}{c}{Uncompressed} & \multicolumn{4}{c}{Resized and Compressed}  \\
 \cmidrule(lr){2-5} \cmidrule(lr){6-9}
 \multicolumn{1}{c}{Acc./AUC\%} & Spec &  PatchFor. & Wang2020 & Grag2021 & Spec &  PatchFor. & Wang2020 & Grag2021 \\
\midrule
ProGAN & \underline{~83.5/~99.2} & \underline{~64.9/~97.6} & \underline{~99.9/100~~} & \underline{~99.9/100~~} & \underline{~49.7/~48.5} & \underline{~50.4/~65.3} & \underline{~99.7/100~~} & \underline{~99.9/100~~} \\
   StyleGAN2 & ~65.3/~72.0 & ~50.2/~88.3 & ~74.0/~97.3 & ~98.1/~99.9 & ~51.8/~50.5 & ~50.8/~73.6 & ~54.8/~85.0 & ~63.3/~94.8 \\
   StyleGAN3 & ~33.8/~ 4.4 & ~50.0/~91.8 & ~58.3/~95.1 & ~91.2/~99.5 & ~52.9/~51.9 & ~50.2/~76.7 & ~54.3/~86.4 & ~58.3/~94.4 \\
      BigGAN & ~73.3/~80.5 & ~52.5/~85.7 & ~66.3/~94.4 & ~95.6/~99.1 & ~52.1/~52.2 & ~50.5/~58.8 & ~55.4/~85.9 & ~79.0/~99.1 \\
        EG3D & ~80.3/~89.6 & ~50.0/~78.4 & ~59.2/~96.7 & ~99.4/100~~ & ~58.9/~60.6 & ~49.8/~81.9 & ~52.1/~85.1 & ~56.8/~96.6 \\
Taming Tran. & ~79.6/~86.6 & ~50.5/~69.4 & ~51.2/~66.5 & ~73.5/~96.6 & ~49.0/~49.1 & ~50.0/~64.1 & ~50.5/~71.0 & ~56.2/~94.3 \\
 DALL·E Mini & ~80.1/~88.1 & ~51.5/~82.2 & ~51.7/~60.6 & ~70.4/~95.6 & ~59.1/~61.9 & ~50.1/~68.7 & ~51.1/~66.2 & ~62.3/~95.4 \\
    DALL·E 2 & ~82.1/~93.3 & ~50.0/~52.5 & ~50.3/~85.8 & ~51.9/~94.9 & ~62.0/~65.0 & ~49.7/~58.4 & ~50.0/~44.8 & ~50.0/~64.4 \\
       GLIDE & ~73.4/~81.9 & ~50.3/~96.6 & ~51.1/~62.6 & ~58.6/~86.4 & ~53.1/~52.5 & ~51.0/~71.5 & ~50.3/~65.9 & ~51.8/~90.0 \\
Latent Diff. & ~72.1/~78.5 & ~51.8/~84.3 & ~51.0/~62.5 & ~58.2/~91.5 & ~47.9/~46.3 & ~50.6/~65.2 & ~50.7/~69.1 & ~52.4/~89.4 \\
Stable Diff. & ~66.8/~74.7 & ~50.8/~85.0 & ~50.9/~65.9 & ~62.1/~92.9 & ~46.5/~44.5 & ~51.1/~77.2 & ~50.7/~72.9 & ~58.1/~93.7 \\
         ADM & ~55.1/~53.3 & ~50.4/~87.1 & ~50.6/~56.3 & ~51.2/~57.4 & ~49.1/~49.1 & ~51.0/~69.1 & ~50.3/~68.1 & ~50.6/~77.2 \\
\midrule
         AVG & ~70.5/~75.2 & ~51.9/~83.2 & ~59.5/~78.6 & ~75.8/~92.8 & ~52.7/~52.7 & ~50.4/~69.2 & ~55.8/~75.0 & ~61.5/~90.8 \\
 \bottomrule
    \end{tabular}
    }
    \caption{Comparative analysis of state-of-the-art techniques. All methods were trained only on ProGAN images, 
    and tested both on uncompressed synthetic data (left) and on resized and compressed data (right).}
    \label{tab:my_label}
\end{table*}

\vspace{2mm}\noindent
{\bf Generalization and robustness.}
First of all, 
we analyze the performance on uncompressed synthetic images in the PNG format, as originated from each model (Table 1, left).
This first experiment aims to highlight that in this situation detection can be much easier, 
because real images, which are always JPEG compressed by a codec embedded in the camera, are characterized by JPEG compression artifacts,
while synthetic images do not embed such traces.
In fact, the AUC performance is almost perfect on ProGAN (seen in training) and remains pretty good also for other architectures.
Even in this favorable case, however, the accuracy is often unsatisfactory,
because the threshold chosen in training does not work well on images of different origin \cite{Gragnaniello2021GAN}.

This is a simple scenario, compared to the situation where both synthetic and real images are compressed and resized as routinely happens on social media platforms.
To simulate such forms of image laundering and to avoid polarization, we follow the procedure used in the IEEE VIP Cup \cite{vipcuplink}. For each image of the test, a crop with random (large) size and position is selected, resized to $200\times 200$ pixels,
and compressed using a random JPEG quality factor from 65 to 100. 
In this challenging condition, Table 1 (right) shows a general reduction of the performance, except for ProGAN (present in training). Again, the performance is acceptable in terms of AUC, but almost random in terms of accuracy. The most difficult diffusion models are DALL·E 2 and ADM, which presented very weak artifacts in our previous analysis.

Then we trained the best performing approach (Grag2021) on images generated using ADM (Table 2, left) and tested on the resized/compressed dataset. First, we observe that an almost perfect detection is achieved not only on ADM but also on Stable Diffusion, coherently with the fact that these architectures share very similar artifacts (Fig.2). The performance on other diffusion models, instead, are not much better than those obtained on GAN generated images. This means that Stable and Latent diffusion models are characterized by different cues compared to ADM and DALL·E 2.  

\vspace{2mm}\noindent
{\bf Fusion and calibration.}
Finally, we carried out a simple experiment where we fuse (simple average) the outputs of the networks trained on both datasets. 
Results are reported in Table 2.
Of course, the performance on GAN generated images improves, while remaining reasonably good on diffusion models, 
however accuracy is still extremely low, due to the unsuitable fixed threshold. 
To improve accuracy we try using a calibration procedure (Platt scaling method \cite{Platt1999probabilistic}) 
under the hypothesis of having just two real images and two synthetic ones for each model. 
Performance greatly improves, but we still cannot reliably detect images that present artifacts significantly different from those seen during training. 

\begin{table}
    \centering
    {\footnotesize
    \begin{tabular}{lccc}
    \toprule 
                                & Trained on  &  \multirow{2}{*}{Fusion} &  \multirow{2}{*}{Calibration} \\
 \multicolumn{1}{c}{Acc./AUC\%} & Latent Diffusion &  &  \\
\midrule
      ProGAN & ~52.0/~78.3 & \underline{~90.2/100~~} & \underline{~97.0/100~~} \\
   StyleGAN2 & ~58.0/~85.0 & ~56.6/~94.6 & ~86.1/~94.6 \\
   StyleGAN3 & ~59.5/~87.6 & ~55.4/~93.9 & ~85.5/~93.9 \\
      BigGAN & ~52.9/~80.6 & ~59.3/~98.5 & ~92.1/~98.5 \\
        EG3D & ~65.4/~91.8 & ~54.4/~97.7 & ~92.3/~97.7 \\
Taming Tran. & ~78.2/~97.3 & ~61.5/~98.2 & ~91.2/~98.2 \\
 DALL·E Mini & ~73.9/~97.3 & ~65.9/~97.7 & ~88.4/~97.7 \\
    DALL·E 2 & ~50.4/~74.2 & ~50.0/~72.5 & ~66.9/~72.5 \\
       GLIDE & ~62.5/~96.2 & ~52.5/~95.9 & ~89.2/~95.9 \\
Latent Diff. & \underline{~97.1/~99.9} & \underline{~84.9/~99.8} & \underline{~96.4/~99.8} \\
Stable Diff. & ~99.7/100~~ & ~92.5/100~~ & ~97.2/100~~ \\
         ADM & ~52.9/~81.9 & ~50.8/~80.6 & ~70.8/~80.6 \\
\midrule
         AVG & ~66.9/~89.2 & ~64.5/~94.1 & ~87.8/~94.1 \\
\bottomrule 
    \end{tabular}
    }
    \caption{Results of Grag2021: trained only on Latent Diffusion (left), fused with Grag2021 trained only on ProGAN (center), with calibration applied after fusion (right).}
    \label{tab:fusion}
\end{table}

\section{Conclusion}
\label{sec:con}
This work addressed the problem of detecting synthetic images generated by diffusion models. 
We first tested whether DM images are characterized by distinctive fingerprints just as GAN images are, obtaining a partial confirmation.
Then we analyzed the performance of some state-of-the-art detectors under different realistic scenarios.
Experimental results vary significantly from model to model, as these are characterized by different forensics cues.
Generalization remains the main hurdle, and detectors trained only on GAN images work poorly on these new images. Including a DM model in training can help to detect images generated by similar diffusion models but results can be unsatisfactory for others. Of course, these are only preliminary results and deeper analyses are necessary to address the problem of DM image detection.

\section{Acknowledgment}

We gratefully acknowledge the support of this research by the Defense Advanced Research Projects Agency (DARPA) under agreement number FA8750-20-2-1004. 
The U.S. Government is authorized to reproduce and distribute reprints for Governmental purposes notwithstanding any copyright notation thereon.
The views and conclusions contained herein are those of the authors and should not be interpreted as necessarily representing the official policies or endorsements, either expressed or implied, of DARPA or the U.S. Government. 

In addition, this work has received funding by the European Union under the Horizon Europe vera.ai project, Grant Agreement number 101070093, and is supported by Google and by the PREMIER project, funded by the Italian Ministry of Education, University, and Research within the PRIN 2017 program.

\balance
\bibliographystyle{IEEEbib}
{\small \bibliography{refs}}

\end{document}